\documentclass{article}

 \PassOptionsToPackage{numbers, compress}{natbib}

\usepackage[final]{neurips_2020}

\usepackage[utf8]{inputenc} 
\usepackage[T1]{fontenc}    
\usepackage{hyperref}       
\usepackage{url}            
\usepackage{booktabs}       
\usepackage{amsfonts}       
\usepackage{nicefrac}       
\usepackage{microtype}      
\usepackage{booktabs}
\usepackage{multirow}
\usepackage{graphicx}
\usepackage{amsmath}
\usepackage{comment}

\title{Learning Explainable Interventions to Mitigate HIV Transmission in Sex Workers Across Five States in India}

\author{
 Raghav Awasthi  \\
 Indraprastha Institute of Information Technology\\
 Delhi 110020\\
 \texttt{raghavaw@iiitd.ac.in}\\
   \And
 Prachi Patel \\
 Catalyst Management Services (CMS) \\
 Bangalore 560094 \\
 \texttt{prachi@cms-india.org } \\
   \AND
 Vineet Joshi \\
 Indraprastha Institute of Information Technology \\
 Delhi 110020 \\
 \texttt{vineet19020@iiitd.ac.in } \\
   \And
Shama Karkal  \\
Swasti Health Catalyst \\
 Bangalore 560094 \\
\texttt{shama@swasti.org} \\
   \And
 Tavpritesh Sethi \\
 Indraprastha Institute of Information Technology \\
 Delhi 110020 \\
\texttt{tavpriteshsethi@iiitd.ac.in } \\
}

\begin{document}

\maketitle
\begin{abstract}

Female sex workers(FSWs) are one of the most vulnerable and stigmatized groups in society. As a result, they often suffer from a lack of quality access to care. Grassroot organizations engaged in improving health services are often faced with the challenge of improving the effectiveness of interventions due to complex influences. This work combines structure learning, discriminative modeling, and grass-root level expertise of designing interventions across five different Indian states to discover the influence of non-obvious factors for improving safe-sex practices in  \textit{FSWs}. A bootstrapped, ensemble-averaged Bayesian Network structure was learned to quantify the factors that could maximize condom usage as revealed from the model. A discriminative model was then constructed using XgBoost and random forest in order to predict condom use behavior The best model achieved 83\% sensitivity, 99\% specificity, and 99\% area under the precision-recall curve for the prediction. Both generative and discriminative modeling approaches revealed that financial literacy training was the primary influence and predictor of condom use in \textit{FSWs}. These insights have led to a currently ongoing field trial for assessing the real-world utility of this approach. Our work highlights the potential of explainable models for transparent discovery and prioritization of anti-HIV interventions in female sex workers in a resource-limited setting.
\end{abstract}

\section{Introduction}
Female sex workers (FSWs) are one of the most vulnerable sections of society and their challenges are multiplied manifold in developing countries. Cultural factors leading to power asymmetry, economic vulnerability, social stigma and lack of access to good healthcare increases their vulnerability for practicing unsafe sex and lack of knowledge of their HIV status. The interventions that could mitigate the spread of infection are well known, e.g., condom usage and testing \cite{das_bringing_2015}. However, the factors that govern these protective behaviors are complex, multi-dimensional and under-studied. With a high burden of HIV and limited resources, developing countries could benefit the most from data-driven evidence and machine learning approaches that can prioritize interventions \cite{wilder_clinical_2020} . In India, community-led organizations (COs) working towards HIV prevention are increasingly using digital tools and collecting massive data, thus opening the possibility of understanding these complex influences. The \textit{Avahan} program \cite{avahan-india_hivaids_avahan_nodate} 
has been working in a phased manner in five Indian states since 2003, with a strong emphasis on improving condom usage behaviors. Work in\textit{ Avahan III} showed the importance of engagement with the \textit{COs} for improving the self-confidence and self-efficacy to procure condoms\cite{mahapatra_sustaining_2020}.  In this study, we decided to further delineate the actionable factors through learning the representation of this complex dataset using structure learning approaches, followed by predictive modeling. The goal of the study was to prioritize the set of actions that would have the strongest influence on higher condom use behavior. We believe that this is the first study combining a  unique dataset, expertise in community engagement and a principled machine-learning approach for transparent discovery and prioritization of interventions for mitigating the spread of HIV in female sex workers.

\section{Methods}
\label{gen_inst}
\subsection{Dataset:}

Survey based data on FSW was collected by professionals social workers under the \textit{Avahan III} program \cite{avahan-india_hivaids_avahan_nodate} in Andhra Pradesh, Karnataka, Maharashtra, Tamil Nadu and Telangana states of India. A pretested, structured questionnaire Outcome Monitoring Survey tool was used to collect information directly on a digital Tablet. In addition to the basic demographic and socio-economic characteristics of the respondents, the detailed questionnaire also captured a variety of other factors including CO membership, sexual behavior, condom use with clients and intimate partners, history of STI (Sexually transmitted infections) during last six months, HIV testing behavior, access to and coverage with gender based violence, social protection, (civic identities and social schemes), financial security services and products (Savings Account, Savings Products, insurance and other financial schemes), reproductive health status as well as financial and  food crisis or insufficiency experienced by respondents. A total of 222 variables from 11016 female sex workers were recorded and considered in this analysis during the period July – September, 2017 (3 years after the programme was initiated). 

\subsection{Structure Learning and Network Inference  :}
Explainability and interpretability are key challenges in \textit{AI} based decision models, especially in the public health settings. The expressive power of Bayesian networks combined with the advances in learning structure directly make them well suited to address complex problems that need transparency. Although the structure learned from observational data and edge-directions \textbf{do not} indicate \textit{causality}, yet these can be used as a \textit{probabilistic reasoning system}\cite{pearl__judea_book_nodate}. 
An ensemble averaged directed acyclic graph (DAG) structure was learned directly from the data. The \textit{DAG} $G$ is defined as a triple, $N = \{X, G, P\}$ over a set of random variables, $X$. It encapsulates the underlying joint probability distribution $P$ that can be factored as the product of probabilities of each node $v$ conditioned upon its parents $pa(v)$. The d-separation criterion\cite{geiger_identifying_1990} allows the directed edges $E \sqsubseteq V \times V$ between the vertices $V$ to represent conditional independence relationships between random variables, hence providing a compact representation of data.
\begin{equation} 
\begin{split}
P(X) = \prod_{v\in V}{P(x_v | x_{pa(v)})}
\end{split}
\end{equation}
where v corresponds to the random variable v and pa(v) are the set of  parent variables.


The structure was learned as a search problem over a constrained space of DAGs and evaluated with the Akaike Information Criterion (\textit{AIC}) as a metric.
Since the DAG representation for a set of conditional independencies is not unique, and only up to Markov Equivalence \cite{pearl__judea_book} an ensemble of such structures learned from bootstrapped samples of data was constructed using a majority voting criterion\cite{koller_probabilistic_2009}. For  conducting inferences and queries, the DAG was parametrized with the posterior probability distributions for all variables.$x_{v_j} \in X$ conditionalized upon an evidence $e=\{e_1, e_2,.. e_m\}$ from a set of variables $X(e)$. Where, the prior marginal distribution $P(x_{v_j})$ is given by the following set of equations




The likelihood was then defined as follows.
\begin{equation} \label{eq4}
\begin{split}
L(x_{v_j}|e) & = P(e|x_{v_j}) \\
& = \sum_{x \in X \setminus \{x_{v_j}\}}  \prod_{i \neq j} P(x_{v_i}| x_{pa(v_i)}) \prod_{x \in X(e))} {E_x} \\
\end{split}
\end{equation}

For each $x_{v_j}\notin X(e)$, where $E_x$ is the evidence function for $x \in X(e)$ and $v_i$ is the node representing $x_{v_i}$. Finally the posterior was inferred through application of the Bayes rule and is always proportional to the estimated Likelihood 
\begin{equation} 
\begin{split}
P(x_{v_j}|e) \propto L(x_{v_j}|e) P(x_{v_j}) 
\end{split}
\end{equation}

where the proportionality constant $P(e)$ can be computed from $P(X,e)$ by summation over $X$ and $P(x_{v_j})$ may be obtained by inference over empty set of evidence.

\subsection{Discriminative Predictive Model for Condom Usage:}

Next we took the discriminative approach to confirm whether the discovered influences from the generative approach had enough predictive power to be useful in the real world settings. Data were partitioned into training (80\%) and testing (20\%) sets with the class imbalance was corrected using the Synthetic Minority Oversampling Technique(SMOTE). Condom usage was defined as the dependent variable and the top predictors were identified.  Different supervised machine learning models like Random-Forest(RF), Support Vector Machine (SVMs), Logistic Regression, XgBoost and AdaBoost were then used to predict condom usage.

\section{Results:}
\subsection{Intervention Score \& Vulnerability}
A Combined Vulnerability Reduction Intervention Coverage Score (CVRICS) was defined and calculated for all  female sex workers in our database . \textit{CVRICS} is a combination of variables like social protection intervention score, financial security intervention score and number of incidents of crisis faced by the \textit{FSW} within the last 6 months and its values range between 0 to 30. These include access to entitlements and social protection schemes, access to formal financial products and incidents of violence and reporting of the same as experienced in the last 6 months before the survey. It was observed that although the observed maximum \textit{CVRICS} score was 21, more than 6000 female sex workers had a score of 7-11 indicating that a majority of \textit{FSWs} were vulnerable either financially, socially or physically. 



\subsection{Inferences from Learned Structure:}

From our 101 bootstrapped ensemble-averaged \textit{DAG} we found that financial empowerment was a key association for condom use. The learned conditional probabilities showed that having at least one financial investment such as RD (Recurrent Deposit) FD (Fixed Deposit) and Chit Fund  \textbf{increased }the probability of using condoms with clients by as much as 6\%. We also observed \textbf{Depression symptoms} to be key influencers on a female sex worker’s confidence to buy a condom on her own. Our model showed that the probability of a \textit{FSW} buying a condom on her own decreases by 14\% if she reported symptoms of depression as measured using CESD short scale. We also identified the determinants for HIV testing and our \textit{DAG} revealed legal education and financial stability as two key influencers for HIV testing. The conditional probability of a \textit{FSW} getting tested in every 6 months, increased by 16\% if she had been trained on legal education. It was also observed that creation of a personalized financial plan increased the probability of HIV testing among the sex workers by 12\%. Although it is intuitively expected that these factors could play a role, our model revealed these to be top influences and also quantified their marginal impact.\\
Government schemes were also observed to strongly influence many key \textit{FSW} related variables. Access to at least one government scheme increases a high level of self-efficacy by 13\%, which in-turn drives condom usage behaviors. Since depression symptoms were influencing condom buying behavior, we observed an association between unwanted pregnancy and depression, although a cause and effect relationship cannot be established. However, this highlights the role of mental vulnerability and unsafe practices, and the need to design holistic interventions.
Finally, although the \textit{DAG} approach identifies potential confounding transparently and does not require confounders to be specified up-front, the structure was evaluated by field experts for possible biases.

\subsection{Predictive model for Condom Usage:}
The marginal probability of \textit{\textbf{FSWs}} using condoms with regular paying clients and non regular paying clients was observed as 88.6\%, 80.07\% respectively. Hence there was certainly a scope of improvement to push this beyond the 90 - 95\% goal, but this also led to class imbalance, which was corrected using \textit{SMOTE} algorithm. Discriminative models were trained on training set and their performance was evaluated on test set \textbf{Table1}. The best performing models were \textbf{Random Forest} and \textbf{XgBoost} for non-regular paying clients  and regular paying clients respectively with accuracy 0.96 in both the cases. The sensitivity of the model was only 0.71 and 0.69 in the best performing models, but the intent of interventions is not usually to screen for positive condom usage class, but the opposite. The specificity of detection of positive class was upwards of 97\% in Random Forest and XgBoost models. Finally, in both the generative \textit{DAG} and discriminative modeling approaches, the presence of financial variables as either top influencers or predictors indicate the potential impact of intervening with increased financial empowerment and literacy in order to instill confidence and self-efficacy in the \textit{FSWs}.
However, this study has several limitations. Firstly, as indicated earlier, it is nearly impossible to derive causal influences from observational data and our study does not claim to do that. However, the usefulness of \textit{DAGs} as probabilistic reasoning systems is well known and little applied in public health settings. This is especially important in sensitive studies where black-box models can lead to serious ethical concerns. Secondly, since the study deals with the highly sensitive issue, the responses of \textit{FSWs} may have some degree of incorrectness. However, we believe that the \textit{FSWs} are not explicitly aware of the structure of the influences, hence it is difficult to confabulate the relationships that were discovered in a data-driven manner. 

\begin{table}[]
\resizebox{\textwidth}{!}{%
\begin{tabular}{|l|l|l|l|l|l|l|l|}
\hline
 & Model & Accuracy & F1 Score & Sensitivity & Specificity & AU-ROC & AU-PRC \\ \hline
\multirow{5}{*}{Condom usage with Non-Regular Paying Clients} & \begin{tabular}[c]{@{}l@{}}Random\\ Forest\end{tabular} & 0.96 {[}0.95,0.97{]} & 0.98 {[}0.97,0.98{]} & 0.76 {[} 0.64, 0.86{]} & 0.97 {[}0.96,0.98{]} & 0.86 {[}0.81,0.91{]} & 0.991 {[}0.987,0.995{]} \\ \cline{2-8} 
 & SVM & 0.94 {[}0.92,0.95{]} & 0.967{[} 0.961,0.975{]} & 0.635{[}0.52,0.74{]} & 0.95 {[}0.94,0.96{]} & 0.79 {[}0.75,0.84{]} & \begin{tabular}[c]{@{}l@{}}0.987\\ {[}0.982,0.992{]}\end{tabular} \\ \cline{2-8} 
 & Logistic Regression & 0.93 {[}0.92,0.96{]} & \begin{tabular}[c]{@{}l@{}}0.96\\ {[}0.95,0.97{]}\end{tabular} & \begin{tabular}[c]{@{}l@{}}0.66 \\ {[}0.54,0.76{]}\end{tabular} & \begin{tabular}[c]{@{}l@{}}0.95\\ {[}0.94,0.96{]}\end{tabular} & \begin{tabular}[c]{@{}l@{}}0.81\\ {[}0.75,0.85{]}\end{tabular} & \begin{tabular}[c]{@{}l@{}}0.98\\ {[}0.98,0.99{]}\end{tabular} \\ \cline{2-8} 
 & AdaBoost & \begin{tabular}[c]{@{}l@{}}0.93\\ {[}0.92,0.95{]}\end{tabular} & \begin{tabular}[c]{@{}l@{}}0.96\\ {[}0.95,0.97{]}\end{tabular} & \begin{tabular}[c]{@{}l@{}}0.71\\ {[}0.61,0.81{]}\end{tabular} & \begin{tabular}[c]{@{}l@{}}0.95\\ {[}0.93,0.96{]}\end{tabular} & \begin{tabular}[c]{@{}l@{}}0.83\\ {[}0.75,0.89{]}\end{tabular} & \begin{tabular}[c]{@{}l@{}}0.99\\ {[}0.98,0.99{]}\end{tabular} \\ \cline{2-8} 
 & XgBoost & \begin{tabular}[c]{@{}l@{}}0.96\\ {[}0.95,0.97{]}\end{tabular} & \begin{tabular}[c]{@{}l@{}}0.979\\ {[}0.972,0.984{]}\end{tabular} & \begin{tabular}[c]{@{}l@{}}0.687\\ {[}0.56,0.82{]}\end{tabular} & \begin{tabular}[c]{@{}l@{}}0.97\\ {[}0.96,0.98{]}\end{tabular} & \begin{tabular}[c]{@{}l@{}}0.83\\ {[}0.77,0.88{]}\end{tabular} & \begin{tabular}[c]{@{}l@{}}0.99\\ {[}0.98,0.993{]}\end{tabular} \\ \hline
\multirow{5}{*}{Condom usage with Regular paying Clients} & \begin{tabular}[c]{@{}l@{}}Random\\ Forest\end{tabular} & 0.97{[}0.96,0.98{]} & \begin{tabular}[c]{@{}l@{}}0.988\\ {[}0.982,0.99{]}\end{tabular} & \begin{tabular}[c]{@{}l@{}}0.82\\ {[}0.71,0.91{]}\end{tabular} & \begin{tabular}[c]{@{}l@{}}0.98\\ {[}0.97,0.99{]}\end{tabular} & \begin{tabular}[c]{@{}l@{}}0.90\\ {[}0.84,0.94{]}\end{tabular} & \begin{tabular}[c]{@{}l@{}}0.993\\ {[}0.991,0.999{]}\end{tabular} \\ \cline{2-8} 
 & SVM & \begin{tabular}[c]{@{}l@{}}0.94\\ {[}0.93,0.96{]}\end{tabular} & \begin{tabular}[c]{@{}l@{}}0.972\\ {[}0.96,0.979{]}\end{tabular} & \begin{tabular}[c]{@{}l@{}}0.77\\ {[}0.61,0.84{]}\end{tabular} & \begin{tabular}[c]{@{}l@{}}0.96\\ {[}0.94,0.97{]}\end{tabular} & \begin{tabular}[c]{@{}l@{}}0.86\\ {[}0.82,0.91{]}\end{tabular} & \begin{tabular}[c]{@{}l@{}}0.991\\ {[}0.98,0.994{]}\end{tabular} \\ \cline{2-8} 
 & Logistic Regression & \begin{tabular}[c]{@{}l@{}}0.94\\ {[}0.92,0.95{]}\end{tabular} & \begin{tabular}[c]{@{}l@{}}0.96\\ {[}0.95,0.97{]}\end{tabular} & \begin{tabular}[c]{@{}l@{}}0.77\\ {[}0.67,0.87{]}\end{tabular} & \begin{tabular}[c]{@{}l@{}}0.95\\ {[}0.93,0.96{]}\end{tabular} & \begin{tabular}[c]{@{}l@{}}0.86\\ {[}0.80,0.91{]}\end{tabular} & \begin{tabular}[c]{@{}l@{}}0.99\\ {[}0.98,0.99{]}\end{tabular} \\ \cline{2-8} 
 & AdaBoost & \begin{tabular}[c]{@{}l@{}}0.94\\ {[}0.92,0.95{]}\end{tabular} & \begin{tabular}[c]{@{}l@{}}0.96\\ {[}0.95,0.97{]}\end{tabular} & \begin{tabular}[c]{@{}l@{}}0.85\\ {[}0.74,0.92{]}\end{tabular} & \begin{tabular}[c]{@{}l@{}}0.94\\ {[}0.93,0.95{]}\end{tabular} & \begin{tabular}[c]{@{}l@{}}0.89\\ {[}0.85,0.93{]}\end{tabular} & \begin{tabular}[c]{@{}l@{}}0.993\\ {[}0.99,0.996{]}\end{tabular} \\ \cline{2-8} 
 & XgBoost & \begin{tabular}[c]{@{}l@{}}0.97\\ {[}0.96,0.98{]}\end{tabular} & \begin{tabular}[c]{@{}l@{}}0.985\\ {[}0.98,0.99{]}\end{tabular} & \begin{tabular}[c]{@{}l@{}}0.87\\ {[}0.78,0.95{]}\end{tabular} & \begin{tabular}[c]{@{}l@{}}097\\ {[}0.96,0.98{]}\end{tabular} & \begin{tabular}[c]{@{}l@{}}0.89\\ {[}0.89,0.96{]}\end{tabular} & \begin{tabular}[c]{@{}l@{}}0.995\\ {[}0.992,0.997{]}\end{tabular} \\ \hline
\end{tabular}
}
\caption{Prediction Model performance with 95\% Confidence Interval }
\label{tab: Table 1}
\end{table}

\label{headings}

\section{Broader Impact}
Sexual vulnerability in any gender is a highly sensitive issue, especially in the developing world, and female sex workers are the most vulnerable. A transparent and explainable approach with expert evaluation is critical to model sensitive issues. We demonstrate the potential of such an approach for learning to improve safe practices and mitigate the risk of HIV transmission in a vulnerable population in India, but our work provides a framework for learning and replicating interventions in any settings across the globe.
\begin{ack}
The work was supported by Google AI for Social Good program.
\end{ack}
\label{headings}
\bibliographystyle{IEEEtran}
\bibliography{Cite.bib}
\end{document}